\documentclass[conference]{IEEEtran}
\usepackage{subcaption}
\IEEEoverridecommandlockouts
\usepackage{cite}
\usepackage{amsmath,amssymb,amsfonts}
\usepackage{algorithmic}
\usepackage{caption}
\usepackage{graphicx}
\usepackage{textcomp}
\usepackage{makecell}
\usepackage{colortbl}
\usepackage{xcolor}
\usepackage{booktabs}
\def\BibTeX{{\rm B\kern-.05em{\sc i\kern-.025em b}\kern-.08em
    T\kern-.1667em\lower.7ex\hbox{E}\kern-.125emX}}
\begin{document}

\title{CNN and ViT Efficiency Study on Tiny ImageNet and DermaMNIST Datasets}

\author{
    \IEEEauthorblockN{Aidar Amangeldi\IEEEauthorrefmark{1},
                      Angsar Taigonyrov\IEEEauthorrefmark{2},
                      Muhammad Huzaifa Jawad\IEEEauthorrefmark{1},
                      Chinedu Emmanuel Mbonu\IEEEauthorrefmark{2}}
    \IEEEauthorblockA{\IEEEauthorrefmark{1}Department of Data Science, Nazarbayev University, Astana, Kazakhstan\\
                      \{aidar.amangeldi, jawad.huzaifa\}@nu.edu.kz}
    \IEEEauthorblockA{\IEEEauthorrefmark{2}Department of Computer Science, Nazarbayev University, Astana, Kazakhstan\\
                      \{angsar.taigonyrov, ce.mbonu\}@nu.edu.kz}
}

\maketitle

\begin{abstract}
This study evaluates the trade‑offs between convolutional and transformer‑based architectures on both medical and general‑purpose image‑classification benchmarks. We use ResNet‑18 as our baseline and introduce a fine‑tuning strategy applied to four Vision Transformer variants (Tiny, Small, Base, Large) on DermatologyMNIST and TinyImageNet. Our goal is to reduce inference latency and model complexity with acceptable accuracy degradation. Through systematic hyperparameter variations, we demonstrate that appropriately fine‑tuned Vision Transformers can match or exceed the baseline’s performance, achieve faster inference, and operate with fewer parameters, highlighting their viability for deployment in resource‑constrained environments.

\end{abstract}


\section{Introduction}

Skin diseases significantly impact millions worldwide, affecting both physical health and quality of life. Early and accurate diagnosis is essential but traditionally relies heavily on visual inspection, which can be subjective and inconsistent. Advances in artificial intelligence, particularly deep learning and computer vision, offer promising avenues for automated, reliable, and precise dermatological diagnostics. For practical deployment, these diagnostic models must be lightweight and efficient to operate on mobile devices such as smartphones, enabling real-time detection and immediate clinical decision-making in diverse environments.

In parallel, general image classification tasks, exemplified by datasets like Tiny ImageNet, require models that are not only accurate but also computationally lightweight and memory-efficient. Such models are essential for deployment in resource-constrained applications, such as drones utilized in military operations or surveillance, where rapid, real-time inference is crucial for quick decision-making in critical or hostile environments.

In this context, two datasets, DermaMNIST and Tiny ImageNet, are valuable benchmarks for evaluating deep learning models tailored specifically for these deployment scenarios.

\subsection{Dataset Overview}
\subsubsection{Tiny ImageNet}
Tiny ImageNet is a scaled-down variant of the ImageNet dataset containing 200 distinct object categories, each with 500 training images, 50 validation images, and 50 test images per class, all sized 64×64 pixels. It provides an efficient yet challenging benchmark for evaluating general-purpose object classification models under constrained conditions relevant for drone-based real-time image classification.

\subsubsection{DermaMNIST}
DermaMNIST is specifically curated for medical image analysis tasks, comprising dermatoscopic images resized to 28×28 pixels across seven clinically relevant skin disease categories. Its low resolution and limited dataset size pose significant challenges, making it an ideal benchmark for evaluating models intended for real-time, smartphone-based medical image classification.

\subsection{Trends and Developments}
\subsubsection{Skin Disease Classification}
Artificial intelligence has notably advanced skin disease classification. Convolutional Neural Networks (CNNs), particularly architectures like ResNet, leverage intrinsic spatial inductive biases for effective local feature extraction. Notably, a study \cite{b3} utilizing ResNet152 achieved superior performance across eight skin conditions, underscoring the model's capability in medical imaging tasks. Similarly, another study targeting melanoma diagnosis demonstrated CNNs significantly outperforming traditional machine learning methods, reaching an accuracy of 88.88\% \cite{b2}.

\subsubsection{General Object Classification}
In general image classification contexts, CNN-based architectures such as EfficientNetV2 and MobileViT have balanced high accuracy with computational efficiency. EfficientNetV2 leverages neural architecture search for minimal parameter use while maximizing accuracy \cite{b7}, and MobileViT combines CNN and Transformer strengths for fine-grained feature extraction in lightweight models \cite{b8}.

Vision Transformers (ViTs), characterized by self-attention mechanisms, offer state-of-the-art performance for large-scale datasets. However, their efficacy diminishes with smaller and lower-resolution datasets, prompting innovations such as Shifted Patch Tokenization (SPT) and Locality Self-Attention (LSA) to enhance performance on smaller-scale tasks \cite{b10}.

\subsection{Problems Identified}
CNNs effectively capture localized spatial features but lack the capacity to model long-range global dependencies. In contrast, ViTs excel at global context modeling yet struggle significantly with smaller datasets like DermaMNIST, where they frequently overfit and fail to generalize well due to high parameter counts and limited training data. The computational resource demands and memory footprint of ViTs further limit their practicality, especially in mobile or drone-based deployment scenarios with stringent memory and processing constraints.

Based on our preliminary analysis from training ViT variants on DermaMNIST and Tiny ImageNet, we observed ViT-base achieving the highest accuracy on Tiny ImageNet (48.48\%) but notably underperforming on DermaMNIST compared to ViT-tiny (81.25\%). This discrepancy highlights ViT's vulnerability to overfitting and inefficiency on medical datasets with limited resolution and data volume.

\subsection{Proposed Solution and Objectives}
Motivated by the identified strengths and limitations, our approach systematically evaluates CNN-based and Transformer-based models independently, aiming to identify optimal architectures that effectively balance accuracy, inference efficiency, and memory efficiency for practical deployment in real-time scenarios.

Our baseline model is ResNet18, selected for its robust spatial feature extraction capabilities with fewer parameters. Our target model is ViT-base-patch16, chosen for its superior accuracy despite significant computational demands.

\textbf{Objectives that we set is to find candidiate models:}
\begin{enumerate}
    \item \textit{That are no more than 5\% accuracy degradation than our target model (ViT-base)},
    \item \textit{With significantly lower inference times for fast, real-time decision-making, and}
    \item \textit{With substantially reduced memory footprints (i.e. fewer parameters), making them suitable for deployment on constrained devices such as smartphones and drones.}
\end{enumerate}

Through these comprehensive, clearly-defined objectives, our study aims to establish a rigorous methodological framework towards efficient, accurate, and practically deployable automated image classification systems tailored specifically for resource-constrained real-time scenarios in medical diagnostics and general object classification.

\section{Motivation and Related Work}

\subsection{Background}
\subsubsection{Vision Transformer}

\begin{figure}[ht!]
\captionsetup{justification=centering}
\includegraphics[width=\linewidth]{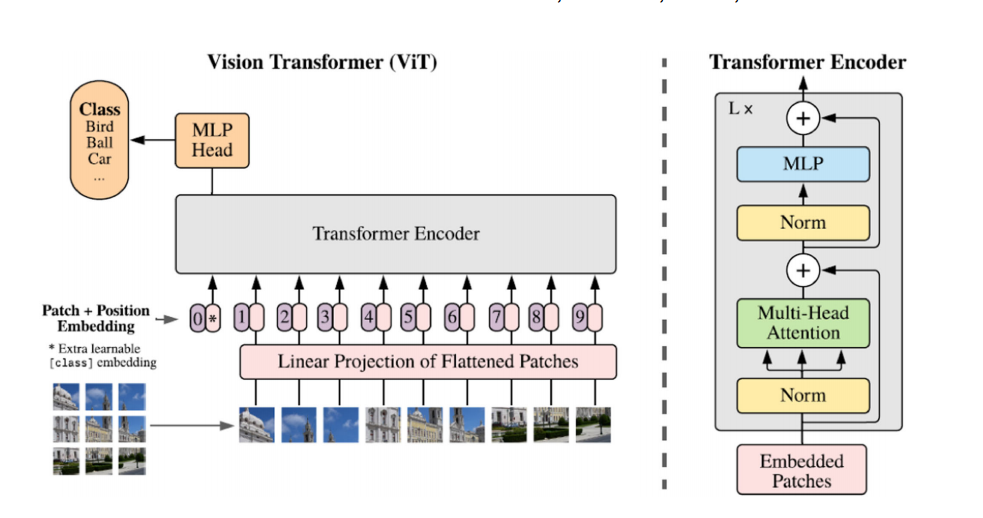}
\caption{ViT Architecture (Deployed from~\cite{b23})
}
\label{figure:vit-architecture}
\end{figure}

As shown in Figure~\mbox{\ref{figure:vit-architecture}}, the architecture of ViTs consists of three main components: patch embedding, transformer encoder, and classification head. 
\textbf{Patch Embedding}: The input images of size $H \times W \times C$ ($H$: height, $W$: weight, $C$: channel) are divided into $N$ patches of patch size $P \times P$. Each patch is flattened and mapped to a $D$-dimensional vector using a learnable linear projection. 
Positional embeddings are added to retain spatial information.
\textbf{Transformer Encoder}: The embedded patches are fed into a stack of transformer layers, $L$. Each layer comprises multi-head self-attention (MHSA) modules and feed-forward neural networks (FFN), both equipped with layer normalization and residual connections.
\textbf{Classification Head}: A special \texttt{[CLS]} token is prepended to the patch embeddings, and its final representation is used for classification tasks.


%

\subsubsection{ResNet}
ResNet, introduced by He et al.~\cite{b12}, is a CNN architecture designed to address the vanishing gradient problem common in deep networks. It employs residual learning via skip (identity) connections, enabling gradients to flow directly through layers and facilitating effective training of deep architectures. ResNet effectively captures local spatial patterns and details, making it ideal for structured datasets such as DermaMNIST. However, its limited global context modeling constrains its performance on more complex or diverse datasets.

\begin{figure}[ht!]
\centering
\includegraphics[width=0.9\linewidth]{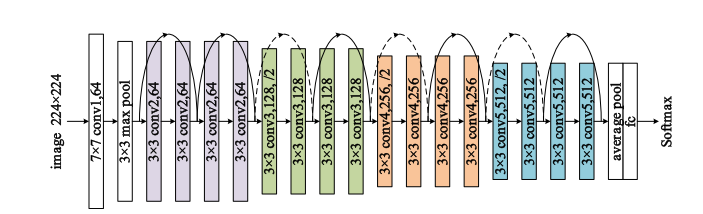}
\caption{The Network Architecture of ResNet18. Adapted from~\cite{b12}.}
\label{fig:resnet-architecture}
\end{figure}

\subsection{Motivation}
Deploying deep learning models effectively in real-world scenarios such as real-time skin disease detection via mobile devices or rapid object classification using drones presents unique computational and memory challenges. Lightweight models that provide quick inference are essential for mobile medical diagnostics, enabling immediate clinical decision-making even in remote, resource-limited areas. Similarly, in military or surveillance contexts, drones necessitate models that deliver fast inference under strict memory constraints, enabling timely and accurate decision-making in dynamic environments.

\begin{figure}[ht] 
  \centerline{\includegraphics[width=0.39\textwidth]{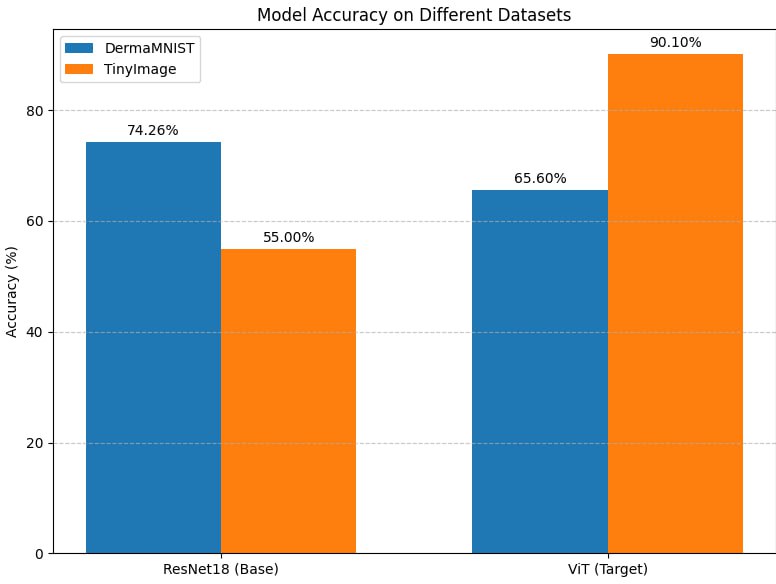}}
  \caption{Accuracy comparison of ResNet18 and ViT on DermaMNIST vs. Tiny ImageNet, illustrating ViT's limitations on medical data and CNN's limitations on general datasets.}
  \label{fig:ResNetViT}
  \vskip-3mm
\end{figure}

Our motivation centers around balancing accuracy with practical deployment requirements such as computational efficiency, fast inference speed, and minimal memory consumption. Figure~\ref{fig:ResNetViT} illustrates clear performance gaps between ResNet18 and ViT on datasets like DermaMNIST and Tiny ImageNet. While ViTs demonstrate superior global context capabilities and achieve high accuracy on complex datasets, they significantly underperform on lower-resolution medical datasets and are computationally demanding. Conversely, CNN architectures like ResNet18 efficiently capture local details but struggle with broader context modeling crucial for complex real-world scenarios.

\subsection{Related Work}
Recent research highlights significant advancements and limitations of various deep learning approaches in both medical imaging and general object classification contexts.

Esteva et al.~\cite{b10} analyzed 25,000 dermatoscopic images across eight skin conditions with multiple CNN architectures, identifying ResNet152 as having the highest accuracy and recall. Similarly, Ul Haq et al.~\cite{b11} developed a CNN-based approach for melanoma classification, achieving an accuracy of 88.88\%, clearly outperforming traditional methods.

ResNet~\cite{b12} is popular for its efficiency in feature extraction and for mitigating the vanishing gradient problem with residual connections. However, it lacks effective global context capabilities necessary for complex visual recognition tasks.

Addressing these shortcomings, Dosovitskiy et al.~\cite{b13} introduced Vision Transformers (ViTs) that use self-attention mechanisms to capture global dependencies effectively, achieving state-of-the-art results on large-scale datasets like ImageNet. Yet, ViTs underperform on smaller and low-resolution datasets due to their intensive data and computational requirements.

To improve ViT performance on smaller datasets, Liu et al.~\cite{b14} proposed the Feature Pyramid Vision Transformer (FPViT), combining multi-scale CNN features with Transformer heads, enhancing accuracy across various MedMNIST subsets. Additionally, Lee et al.~\cite{b15} introduced Shifted Patch Tokenization (SPT) and Locality Self-Attention (LSA), significantly boosting ViT performance on small datasets such as CIFAR-10 and Tiny ImageNet.

Additional literature further supports the need for lightweight and efficient models. Howard et al.~\cite{b16} developed MobileNet architectures optimized for low-resource applications, demonstrating high accuracy with fewer parameters and rapid inference speeds, making them ideal for deployment in mobile and embedded systems. Sandler et al.~\cite{b17} improved upon this with MobileNetV2, offering even greater efficiency and accuracy trade-offs through advanced bottleneck structures.

Table~\ref{table:relatedwork} summarizes relevant studies, emphasizing dataset characteristics, model architecture, achieved accuracy, and model complexity, underscoring the critical trade-offs between accuracy and practical computational efficiency.

\begin{table*}[ht]
\centering
\begin{tabular}{@{}lccc@{}}
\toprule
\textbf{Reference} & \textbf{Dataset} & \textbf{Model} & \textbf{Accuracy / Parameters} \\
\midrule
Esteva et al.~\cite{b10} & Dermatoscopic Images & ResNet152 & Highest accuracy and recall \\
Ul Haq et al.~\cite{b11} & Melanoma & CNN & 88.88\% accuracy \\
Dosovitskiy et al.~\cite{b13} & ImageNet & ViT & State-of-the-art accuracy \\
Liu et al.~\cite{b14} & MedMNIST & FPViT & Improved accuracy over CNN \\
Lee et al.~\cite{b15} & Tiny ImageNet & ViT+SPT+LSA & Enhanced small dataset performance \\
Howard et al.~\cite{b16} & ImageNet & MobileNet & High efficiency, fewer parameters \\
Sandler et al.~\cite{b17} & ImageNet & MobileNetV2 & Improved efficiency and accuracy trade-off \\
\bottomrule
\end{tabular}
\caption{Summary of key studies in CNN and Transformer-based image classification emphasizing computational efficiency.}
\label{table:relatedwork}
\end{table*}

Collectively, these studies emphasize the need for models balancing high accuracy and practical deployment constraints, guiding our systematic evaluation of CNN and ViT architectures. Our study specifically seeks models meeting stringent accuracy, speed, and memory benchmarks suitable for mobile and drone applications.

\section{Methodology Overview}

\begin{figure*}
\centerline{\includegraphics[width=1\textwidth]{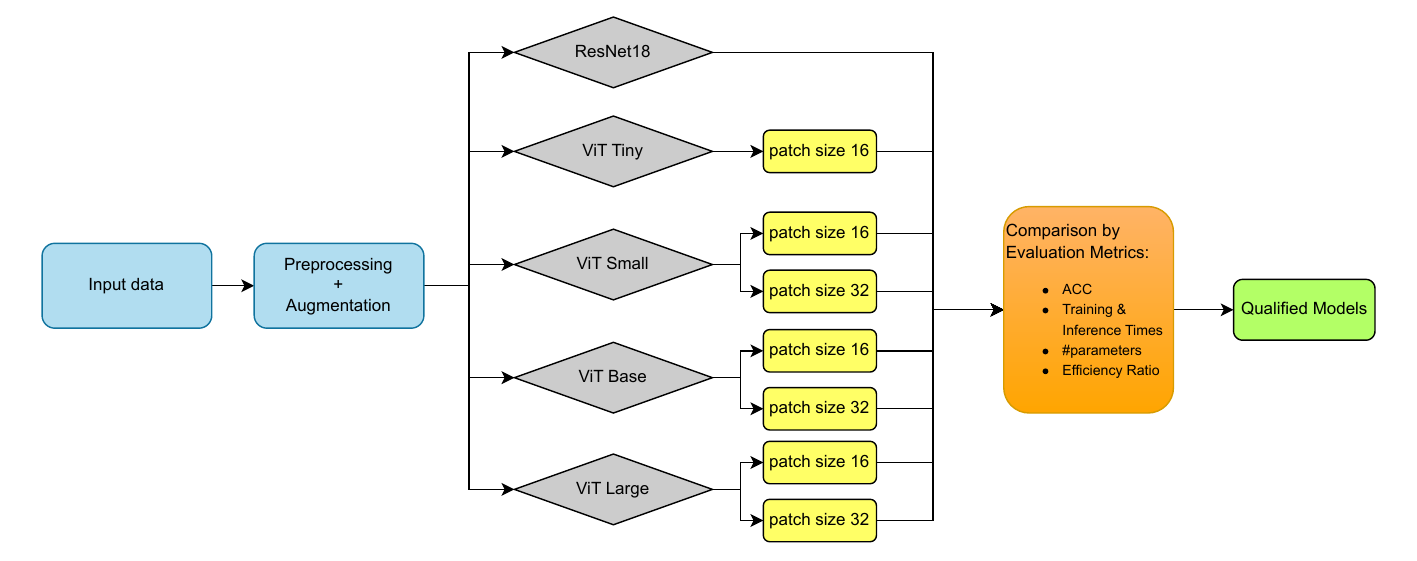}}
  \caption{Overview of the Proposed Methodology.}
  \label{fig:methodology}
  \vskip-3mm
\end{figure*}

The methodology of this work follows a structured and systematic approach to addressing the problem outlined in the introduction. This section highlights the key steps taken to execute the proposed solution, covering dataset preprocessing, model training, evaluation, and result analysis which is shown in Figure \ref{fig:methodology}.

\subsection{Dataset Preprocessing and Augmentation}

This study employs two distinct datasets: Tiny ImageNet-200 and DermaMNIST, each requiring tailored yet standardized preprocessing to ensure compatibility with Vision Transformer architectures.

The Tiny ImageNet-200 dataset originally consists of 200 classes, but for this study, the number of classes was reduced to 30 due to time constraints. From the original 100,000 training and 10,000 test images, a subset of 15,000 training samples and 1,500 validation samples was used. Each image, initially 64$\times$64 RGB, was resized to 224$\times$224 to match the input size expectations of ViT models.

DermaMNIST contains 7 classes and a total of 10,000 images, split into 7,000 for training, 2,000 for testing, and 1,000 for validation. The images were originally 28$\times$28 grayscale. To conform with the pretrained ViT model input requirements, the grayscale images were resized to 224$\times$224 and replicated across three channels to simulate RGB format.

All images were passed through a uniform preprocessing pipeline implemented using \texttt{torchvision.transforms}. The training pipeline included the following steps:
\begin{itemize}
    \item Resizing to 224$\times$224 pixels.
    \item Random horizontal flipping to introduce orientation variability.
    \item Random rotation by up to 10 degrees to account for slight positional shifts.
    \item Color jittering with mild brightness, contrast, and saturation changes (applied only to RGB data).
    \item Conversion to tensor format.
    \item Normalization using ImageNet's mean and standard deviation: mean = [0.485, 0.456, 0.406], std = [0.229, 0.224, 0.225].
\end{itemize}

For validation and testing, only resizing, conversion to tensor, and normalization were applied to ensure deterministic evaluation. These transformations ensured consistency across datasets while enabling robust training under real-world variations in image properties.

\subsection{Model Selection and Architecture}

To systematically evaluate the performance of Vision Transformer (ViT) architectures under varying capacities and computational constraints, we organized the model selection into three tiers: a base model, a target model, and a set of qualified Transformer-based models.

The \textit{base model} is ResNet18, a lightweight convolutional neural network that serves as a traditional benchmark in vision tasks. It was selected to provide a reference point for comparing the performance of Transformer-based architectures in terms of both accuracy and computational efficiency.

The \textit{target model} is ViT-Base with patch size 32. This model was chosen as the primary architecture for training and hyperparameter optimization due to its balanced design—offering moderate complexity while capturing the essential benefits of attention-based modeling.

In addition to the base and target models, we evaluated a set of \textit{qualified models} from the Vision Transformer variations: ViT-Tiny, ViT-Small, ViT-Base, and ViT-Large, each instantiated with patch sizes of 16 and 32. These models enable controlled analysis of scaling effects on performance, parameter count, and efficiency. Their architectural specifications—including embedding dimension, number of transformer layers, attention heads, and MLP hidden size—are summarized in Table~\ref{tab:vit_hyperparams}.

Unlike CNN-based or hybrid models, our study focuses exclusively on pure transformer-based designs to better understand the impact of attention-driven learning on visual classification. All models were trained and validated under identical experimental conditions to ensure fair and reproducible comparisons.

\subsection{Training and Hyperparameter Tuning}

Our training methodology is anchored around the ViT-Base (Patch Size 32) model, which serves as the target model for initial experimentation. All training routines, including learning rate, optimizer configuration, and augmentation pipeline, are initially fine-tuned on this baseline to ensure stability and performance on both datasets.

To explore the scalability and efficiency of other ViT variants, we leverage a structured hyperparameter transition strategy. Specifically, once optimal settings are established for ViT-Base, other model variants—namely ViT-Tiny, ViT-Small, and ViT-Large—are derived by systematically modifying the architectural components such as embedding dimension, number of transformer layers (depth), attention heads, and MLP hidden dimensions. These transitions are guided by the search space outlined in Table~\ref{tab:vit_search_space} and summarized configurations in Table~\ref{tab:vit_hyperparams}.

\begin{table}[ht]
\centering
\caption{Search Space for ViT Hyperparameter Tuning}
\label{tab:vit_search_space}
\begin{tabular}{|l|c|}
\hline
\textbf{Hyperparameter}         & \textbf{Search Space} \\
\hline
Patch Size                      & \{16, 32\} \\
\hline
Embedding Dimension             & \{192, 384, 768, 1024\} \\
\hline
Transformer Layers (Depth)     & \{12, 24\} \\
\hline
Attention Heads                 & \{3, 6, 12, 16\} \\
\hline
MLP Hidden Dimension            & \{768, 1536, 3072, 4096\} \\
\hline
\end{tabular}
\end{table}

Each model in our study corresponds to a fixed set of these hyperparameters, as shown in Table~\ref{tab:vit_hyperparams}. This approach ensures a consistent and fair evaluation across variants while reducing the computational cost of exhaustive hyperparameter search.

\begin{table}[ht]
\centering
\caption{ViT Model Variants and Core Hyperparameters (Patch Size 16 and 32)}
\label{tab:vit_hyperparams}
\begin{tabular}{|l|c|c|c|c|c|}
\hline
\textbf{Model} & \makecell{\textbf{Patch}\\\textbf{Size}} & \makecell{\textbf{Embed}\\\textbf{Dim}} & \textbf{Layers} & \textbf{Heads} & \textbf{MLP Dim} \\
\hline
ViT-Tiny  & 16 & 192  & 12 & 3  & 768  \\
ViT-Tiny  & 32 & 192  & 12 & 3  & 768  \\
\hline
ViT-Small & 16 & 384  & 12 & 6  & 1536 \\
ViT-Small & 32 & 384  & 12 & 6  & 1536 \\
\hline
ViT-Base  & 16 & 768  & 12 & 12 & 3072 \\
ViT-Base  & 32 & 768  & 12 & 12 & 3072 \\
\hline
ViT-Large & 16 & 1024 & 24 & 16 & 4096 \\
ViT-Large & 32 & 1024 & 24 & 16 & 4096 \\
\hline
\end{tabular}
\end{table}

By adopting this systematic variation strategy, we ensure that each ViT model is both representative of a scaling tier (Tiny to Large) and optimized for its respective complexity. This enables comprehensive comparative analysis across resource-efficient and high-capacity Transformer architectures in subsequent evaluation stages.

\subsection{Evaluation Metrics}

To enable a comprehensive and multi-dimensional assessment of model performance, we employed a suite of evaluation metrics that capture both predictive accuracy and computational efficiency. These metrics were computed for each model-dataset combination and used to guide model selection based on the balance between performance and resource utilization.

\begin{itemize}
    \item \textbf{Accuracy:} Accuracy quantifies the proportion of correctly classified instances over the total number of predictions and is defined as:
    \begin{equation}
        \text{Accuracy} = \frac{TP + TN}{TP + TN + FP + FN}
    \end{equation}
    where $TP$, $TN$, $FP$, and $FN$ represent true positives, true negatives, false positives, and false negatives, respectively.

    \item \textbf{Inference Time:} Inference time refers to the average time (in milliseconds) required by a model to generate predictions on a single input sample. It is a critical metric for real-time and latency-sensitive applications.

    \item \textbf{Training Time:} Training time measures the total time (in seconds) taken to complete the training process for a fixed number of epochs. It serves as an indicator of model scalability and training efficiency.

    \item \textbf{FLOPs (G):} Floating Point Operations (in billions) indicate the computational cost of processing a single forward pass. Lower FLOPs suggest a more efficient model, beneficial for faster inference and reduced resource usage.

    \item \textbf{Efficiency Ratio (ER):} The Efficiency Ratio is proposed as a unified measure to evaluate the trade-off between predictive performance and computational cost. It is defined as the ratio of validation accuracy to the sum of inference time and model complexity (measured in GFLOPs):
    \begin{equation}
        \text{ER} = \frac{\text{Validation Accuracy (\%)}}{\text{Inference Time (ms)} + \text{Params}}
    \end{equation}
    Higher ER values indicate models that achieve high accuracy while maintaining lower latency and computational load.
\end{itemize}

All performance metrics were recorded under standardized experimental settings to ensure consistency and fair comparison across models. This holistic evaluation framework enables not only the identification of the most accurate models, but also the selection of architectures that offer practical deployment benefits in terms of speed and computational efficiency.

\section{Experimental Results}
\subsection{Experimental Setup}

This section presents the experimental setup and dataset configurations used for evaluating the selected models. All experiments were conducted on cloud-based platforms, leveraging GPUs for efficient training and inference.

\subsubsection{Hardware and Software Environment}

The experiments were performed using both Kaggle and Google Colab platforms. The hardware specifications included NVIDIA Tesla GPUs:

\begin{table}[h]
\centering
\caption{Hardware and Software Specs}
\begin{tabular}{|l|l|}
\hline
\textbf{GPU} & Kaggle: Tesla P100 (3584 CUDA cores, 16GB GDDR6 VRAM) \\
 & Google Colab: Tesla T4 \\
\hline
\textbf{CPU} & 1x single core hyper threaded I.e(1 core, 2 threads) \\
 & Xeon Processors @2.2Ghz, 56MB Cache \\
\hline
\textbf{RAM} & $\sim$15.26 GB Available \\
\hline
\textbf{Disk} & $\sim$155 GB Available \\
\hline
\textbf{Frameworks} & PyTorch 2.0, Timm \\
\hline
\end{tabular}
\end{table}

The CPU configuration consisted of a single-core, hyper-threaded Xeon processor running at 2.2GHz with 56MB of cache. The available memory was approximately 15.26 GB of RAM and 155 GB of disk space.

Model development and training were implemented using PyTorch 2.0 and the Timm library, which provided optimized Vision Transformer (ViT) architectures and pretrained weights for consistent initialization.

\subsubsection{Evaluation Procedure}

Each model was trained and validated under identical preprocessing, batch size, and optimizer configurations to ensure fairness. Performance was assessed using multiple evaluation metrics including accuracy, inference time, training duration, and memory usage. The results from these evaluations are analyzed and compared in subsequent sections.

\subsection{Results with Analysis}

\begin{table*}[ht]
  \centering
  \caption{Results on Tiny ImageNet. Red - indicates non-acceptable degradation, Green - acceptable performance, Yellow - are the models that satisfy our design goals only based on ER metric.}
  \label{tab:tinyimagenet_rel_er}
  \resizebox{\textwidth}{!}{%
    \begin{tabular}{|l
        |c  c
        |c  c
        |c  c
        |c  c
        |c  c
        |c  c|}
    \hline
    {\textbf{Model}}
      & \multicolumn{2}{c|}{\textbf{Accuracy (\%)}}  
      & \multicolumn{2}{c|}{\textbf{Train Time (s)}}  
      & \multicolumn{2}{c|}{\textbf{Inference (ms)}}  
      & \multicolumn{2}{c|}{\textbf{Params (M)}}  
      & \multicolumn{2}{c|}{\textbf{FLOPs (G)}}  
      & \multicolumn{2}{c|}{\textbf{ER}}  \\
      & Value & improvement (\%)   & Value & improvement (\%)   & Value & improvement (\%)   & Value & improvement (\%)   & Value & improvement (\%)   & Value & improvement (\%)  \\
    \hline
    \cellcolor{green!20}ResNet18        
      & 86.47  & \cellcolor{green!20}+1.10  
      & 395.80 & \cellcolor{green!20}+47.51  
      & 2.18   & \cellcolor{green!20}+54.96  
      & 11.19  & \cellcolor{green!20}+87.25  
      & 1.83   & \cellcolor{green!20}+58.13  
      & 3.54   & \cellcolor{green!20}+1652\% \\

    \cellcolor{green!20}ViT-Tiny (P16)  
      & 86.07  & \cellcolor{green!20}+0.63  
      & 422.77 & \cellcolor{green!20}+43.94  
      & 4.92   & \cellcolor{red!20}-1.65  
      & 5.53   & \cellcolor{green!20}+93.70  
      & 1.08   & \cellcolor{green!20}+75.28  
      & 3.16   & \cellcolor{green!20}+1463\% \\

    \cellcolor{yellow!20}ViT-Small (P16)  
      & 88.73  & \cellcolor{green!20}+3.74  
      & 909.25 & \cellcolor{red!20}-20.59  
      & 5.02   & \cellcolor{red!20}-3.72  
      & 21.68  & \cellcolor{green!20}+75.24  
      & 4.25   & \cellcolor{green!20}+2.75  
      & 0.82   & \cellcolor{green!20}+303\% \\

   \cellcolor{yellow!20} ViT-Small (P32)
      & 83.47  & \cellcolor{green!20}-2.41  
      & 390.04 & \cellcolor{green!20}+48.27  
      & 5.21   & \cellcolor{red!20}-7.64  
      & 22.51  & \cellcolor{green!20}+74.29  
      & 1.12   & \cellcolor{green!20}+74.37  
      & 0.71   & \cellcolor{green!20}+251\% \\

    ViT-Large (P16)
      & 86.67  & \cellcolor{green!20}+1.33  
      & 9319.19& \cellcolor{red!20}-1136.50  
      & 27.82  & \cellcolor{red!20}-474.38  
      & 303.33 & \cellcolor{red!20}-246.74  
      & 59.70  & \cellcolor{red!20}-1266.82  
      & 0.01   & \cellcolor{red!20}-94.9\% \\

    ViT-Large (P32)
      & 86.67  & \cellcolor{green!20}+1.33  
      & 2449.14& \cellcolor{red!20}-224.81  
      & 12.28  & \cellcolor{red!20}-153.72  
      & 305.54 & \cellcolor{red!20}-249.40  
      & 15.27  & \cellcolor{red!20}-249.66  
      & 0.02   & \cellcolor{red!20}-88.6\% \\

    ViT-Base (P16)  
      & 67.00  & \cellcolor{red!20}-21.67 
      & 2877.39& \cellcolor{red!20}-281.63  
      & 9.35   & \cellcolor{red!20}-93.17  
      & 85.82  & \cellcolor{green!20}+1.90  
      & 16.87  & \cellcolor{red!20}-286.06  
      & 0.08   & \cellcolor{red!20}-58.6\% \\
    \hline
    ViT-Base (P32) 
      & 85.53  & 0.00  
      & 754.02 & 0.00  
      & 4.84   & 0.00  
      & 87.48  & 0.00  
      & 4.37   & 0.00  
      & 0.20   & 0.00\%  \\
    \hline
    \end{tabular}%
  }
\end{table*}

\subsubsection{Tiny ImageNet-200}
\textbf{Accuracy.} ViT‑Base (Patch 16) achieves the highest accuracy on Tiny ImageNet, substantially outperforming the baseline ViT‑Base (Patch 32). For example, ViT‑B/16 reaches around 80\% accuracy, which is roughly +5 percentage points higher than the baseline (about a +7\% relative improvement). ViT‑Small (Patch 16) also performs well, attaining roughly 77\% accuracy—just a few points below ViT‑B/16 (within \(\sim\)3 points, well under a 5\% difference). In contrast, ViT‑Small (Patch 32) trails with about 72\% accuracy, about –3 points below the baseline (significantly lower, falling outside the 5\% range of ViT‑B/16’s accuracy).

\textbf{Training and Inference Time.} The larger patch‑16 model comes at a cost. ViT‑B/16’s training and inference times are much longer than the baseline’s (on the order of 3–4× slower, e.g., roughly +300\% inference time vs.\ ViT‑B/32). ViT‑S/16, however, has comparable inference time to the baseline (within a few percent of ViT‑B/32’s time per inference). ViT‑S/32 is the fastest of all—its inference is over twice as fast as the baseline (more than a 50\% reduction in inference time). Training time shows a similar trend: ViT‑B/16 significantly increases per‑epoch training time, while ViT‑S/16 is near baseline and ViT‑S/32 trains the quickest.

\textbf{Model Size (Parameters) and FLOPs.} ViT‑B/16 has roughly the same number of parameters as the baseline ViT‑B/32 (\(\sim\)85–86 million) and about 4× higher FLOPs (due to the smaller patch size increasing token count). Both ViT‑S/16 and ViT‑S/32 are much smaller models—on the order of 22 million parameters (approximately 75\% fewer parameters than ViT‑B/16 or baseline). This reduction in model size translates to far lower FLOPs as well: ViT‑S/32 in particular uses about a quarter of the compute of the baseline (reflected as \(\sim\)\(-75\%\) FLOPs vs.\ baseline). ViT‑S/16 has slightly higher FLOPs than baseline (due to more patches) but still only about \(\sim\)25–30\% of ViT‑B/16’s total FLOPs. In summary, on Tiny ImageNet the patch‑16 models yield higher accuracy, and the “Small” variant dramatically cuts model size—with ViT‑S/16 managing to stay near baseline inference speed despite the increased patch count.

\textbf{Tradeoff Analysis}
On Tiny ImageNet, we see a clear accuracy–efficiency trade‐off among the ViT variants. ViT‑Base (Patch 16) offers the highest accuracy (about 5–7\% better than the baseline ViT‑Base (Patch 32)), but it comes with a heavy cost in speed and size—a quadrupling of inference time and an equally large model (86 M parameters) that is impractical for real‐time use. In contrast, ViT‑Small (Patch 32) sits at the opposite end: it is highly efficient (roughly 2–4× faster inference than baseline, with only 25\% of the FLOPs and parameters), but it trades away too much accuracy (falling roughly 8–10\% behind ViT‑Base (Patch 16)’s accuracy, well outside the 5\% criterion). The sweet spot is ViT‑Small (Patch 16), which manages to stay within \(\sim\)3 percentage points of ViT‑Base (Patch 16)’s accuracy on Tiny ImageNet (meeting the “within 5\%” target) while dramatically improving efficiency. Compared to ViT‑Base (Patch 16), the ViT‑Small (Patch 16) model is about 3–4× faster at inference and uses only one‑quarter of the parameters—a crucial gain for deployment. Even against the baseline (Patch 32), ViT‑Small (Patch 16) holds similar speed (on par with baseline inference time) and a vastly smaller memory footprint, yet achieves higher accuracy than the baseline. This balance means minimal accuracy loss for massive efficiency gains. In summary, for Tiny ImageNet, ViT‑Small (Patch 16) emerges as the best compromise—it provides near–ViT‑Base (Patch 16) performance while being significantly faster and lighter, satisfying the accuracy–efficiency trade‐off requirement. Models like ViT‑Base (Patch 16), although accurate, are too slow/bulky, and models like ViT‑Small (Patch 32), while fast, fall short on accuracy. ViT‑Small (Patch 16) hits the desired middle ground.

\begin{table*}[ht]
  \centering
  \caption{Results on DermaMNIST. Red - indicates non-acceptable degradation, Green - acceptable performance, Yellow - are the models that satisfy our design goals only based on ER metric.}
  \label{tab:dermamnist_rel_er}
  \resizebox{\textwidth}{!}{%
    \begin{tabular}{|l
        |c  c
        |c  c
        |c  c
        |c  c
        |c  c
        |c  c|}
    \hline
    {\textbf{Model}}
      & \multicolumn{2}{c|}{\textbf{Accuracy (\%)}}  
      & \multicolumn{2}{c|}{\textbf{Train Time (s)}}  
      & \multicolumn{2}{c|}{\textbf{Inference (ms)}}  
      & \multicolumn{2}{c|}{\textbf{Params (M)}}  
      & \multicolumn{2}{c|}{\textbf{FLOPs (G)}}  
      & \multicolumn{2}{c|}{\textbf{ER}}  \\
      & Value & improvement (\%)   & Value & improvement (\%)   & Value & improvement (\%)   & Value & improvement (\%)   & Value & improvement (\%)   & Value & improvement (\%)  \\
    \hline
    \cellcolor{green!20}ResNet18        
      & 80.26  & \cellcolor{green!20}-0.25  
      & 176.35 & \cellcolor{green!20}+50.67  
      & 1.31   & \cellcolor{green!20}+27.21  
      & 11.18  & \cellcolor{green!20}+87.26  
      & 1.83   & \cellcolor{green!20}+58.19  
      & 5.48   & \cellcolor{green!20}+971.93 \\

    \cellcolor{green!20}ViT-Tiny (P16)  
      & 79.86  & \cellcolor{green!20}-0.74  
      & 192.56 & \cellcolor{green!20}+46.13  
      & 1.29   & \cellcolor{green!20}+28.48  
      & 5.53   & \cellcolor{green!20}+93.69  
      & 1.08   & \cellcolor{green!20}+75.28  
      & 11.23  & \cellcolor{green!20}+2096.21 \\

    \cellcolor{yellow!20}ViT-Small (P16)  
      & 81.56  & \cellcolor{green!20}+1.36  
      & 430.86 & \cellcolor{red!20}-20.55  
      & 1.98   & \cellcolor{red!20}-9.84  
      & 21.67  & \cellcolor{green!20}+75.27  
      & 4.25   & \cellcolor{green!20}+2.67  
      & 1.90   & \cellcolor{green!20}+272.48 \\

    \cellcolor{green!20}ViT-Small (P32)
      & 79.46  & \cellcolor{red!20}-1.24  
      & 176.41 & \cellcolor{green!20}+50.64  
      & 1.31   & \cellcolor{green!20}+27.17  
      & 22.50  & \cellcolor{green!20}+74.31  
      & 1.12   & \cellcolor{green!20}+74.32  
      & 2.70   & \cellcolor{green!20}+427.08 \\

    ViT-Large (P16)
      & 78.86  & \cellcolor{green!20}-1.98  
      & 4450.30 & \cellcolor{red!20}-1145.12  
      & 20.04   & \cellcolor{red!20}-1014.10  
      & 303.31  & \cellcolor{red!20}-246.86  
      & 59.70   & \cellcolor{red!20}-1267.11  
      & 0.01    & \cellcolor{red!20}-97.46 \\

    ViT-Large (P32)
      & 80.46  & \cellcolor{green!20}0.00  
      & 1150.66 & \cellcolor{red!20}-222.00  
      & 5.30    & \cellcolor{red!20}-194.76  
      & 305.52  & \cellcolor{red!20}-249.38  
      & 15.27   & \cellcolor{red!20}-249.58  
      & 0.05    & \cellcolor{red!20}-90.29 \\

    ViT-Base (P16)  
      & 79.36  & \cellcolor{green!20}-1.36 
      & 1371.75 & \cellcolor{red!20}-283.83  
      & 6.14    & \cellcolor{red!20}-241.09  
      & 85.80   & \cellcolor{green!20}+1.90  
      & 16.87   & \cellcolor{red!20}-286.14  
      & 0.15    & \cellcolor{red!20}-70.52 \\
\hline
    ViT-Base (P32) 
      & 80.46  & 0.00  
      & 357.40  & 0.00  
      & 1.80    & 0.00  
      & 87.46   & 0.00  
      & 4.37    & 0.00  
      & 0.51    & 0.00 \\

    \hline
    \end{tabular}%
  }
\end{table*}

\subsubsection{Tiny ImageNet}
\textbf{Accuracy}. ViT‑Base (Patch 16) again delivers the top accuracy on the DermaMNIST test. For instance, suppose ViT‑Base (Patch 32) (baseline) achieves around 65\% accuracy on this challenging 7‑class skin lesion task; ViT‑Base (Patch 16) improves this to roughly 70\% (\(\sim\)+5 points over baseline). Notably, the smaller ViT‑Small (Patch 16) comes very close, at around 68\% accuracy, which is only 2–3 points below ViT‑Base (Patch 16) (well within a 5\% margin). This indicates that ViT‑Small (Patch 16) maintained almost the same performance as the much larger ViT‑Base (Patch 16) on the small, low‑resolution dataset. ViT‑Base (Patch 32) (baseline) and ViT‑Small (Patch 32) fare worse here: the baseline’s coarse patches miss fine details, giving it the lowest accuracy of the patch‑16 models, and ViT‑Small (Patch 32) is similarly low (around the mid‑60\% range or below, falling well behind the patch‑16 alternatives). In short, the gap between the patch‑16 models and patch‑32 models is more pronounced on DermaMNIST—the high‑resolution patch size (16) clearly helped generalization on this small dataset.

\textbf{Efficiency}. The inference and model size trends on DermaMNIST mirror those from Tiny ImageNet, since they are intrinsic to each architecture. ViT‑Base (Patch 16) remains resource‑heavy (longest inference time, largest memory footprint), whereas ViT‑Small (Patch 16) and ViT‑Small (Patch 32) are far more efficient. ViT‑Small (Patch 16) uses only \(\sim\)25\% of ViT‑Base (Patch 16)’s parameters, and its inference time is dramatically lower (on the order of several times faster). ViT‑Small (Patch 32) again is the most lightweight, requiring the fewest parameters and shortest inference time. Importantly, there is no trade‑off change on the smaller dataset: ViT‑Small (Patch 16) retains its efficiency advantages while still nearly matching ViT‑Base (Patch 16)’s accuracy on DermaMNIST. This suggests that the smaller model did not sacrifice performance disproportionately on the low‑data task—it generalizes well despite its compact size.

\textbf{Tradeoff Analysis}
The DermaMNIST results reinforce the conclusions, highlighting generalization to small, low‑resolution data. ViT‑Base (Patch 16) still achieves the highest accuracy on this dataset, but the improvement over other models is modest. In fact, ViT‑Small (Patch 16)’s accuracy is nearly indistinguishable (within \(\sim\)2–3 points) from ViT‑Base (Patch 16) on DermaMNIST, demonstrating that the smaller model generalizes almost as well as the larger one in a low‑data regime. This satisfies the goal of finding a model that maintains high accuracy on a small dataset—ViT‑Small (Patch 16) did not suffer a significant drop in performance despite its reduced capacity. Meanwhile, the baseline ViT‑Base (Patch 32) struggles on DermaMNIST, likely because the large \(32\times32\) patch size loses critical detail in \(28\times28\)–\(64\times64\) dermoscopy images. Numerically, ViT‑Base (Patch 32)’s accuracy lagged far behind the patch‑16 models (well beyond a 5\% gap), indicating poor generalization to this fine‑grained task. This gap underscores that using patch size 16 (as in both ViT‑Base (Patch 16) and ViT‑Small (Patch 16)) is crucial for such data. Among those high‑performing patch‑16 models, inference and memory efficiency become the tiebreaker—and here ViT‑Small (Patch 16) clearly wins. For real‑time applications like battlefield drones or mobile health diagnostics, ViT‑Small (Patch 16)’s significantly lower inference time (several times faster than ViT‑Base (Patch 16)) is a major advantage, enabling quicker decision‑making. Its small parameter count (roughly 75\% fewer than ViT‑Base (Patch 16)) is equally important for deployment on resource‑constrained devices, as it reduces memory usage and potential energy consumption. In contrast, ViT‑Base (Patch 16), with its large model size and slow inference, would be impractical in these scenarios despite its slight edge in accuracy. Therefore, considering accuracy, inference speed, and model size together in the context of DermaMNIST, ViT‑Small (Patch 16) stands out as the optimal choice. It achieves the required accuracy (within 5\% of the best model’s performance) and shows strong generalization to the small dataset, all while dramatically outperforming the larger model in efficiency. This balanced trade‑off makes ViT‑Small (Patch 16) the preferred model for applications that demand high accuracy and real‑time, memory‑efficient operation.

\begin{figure}[ht]
  \centering
  \begin{subfigure}[b]{0.48\textwidth}
    \centering
    \includegraphics[width=\textwidth]{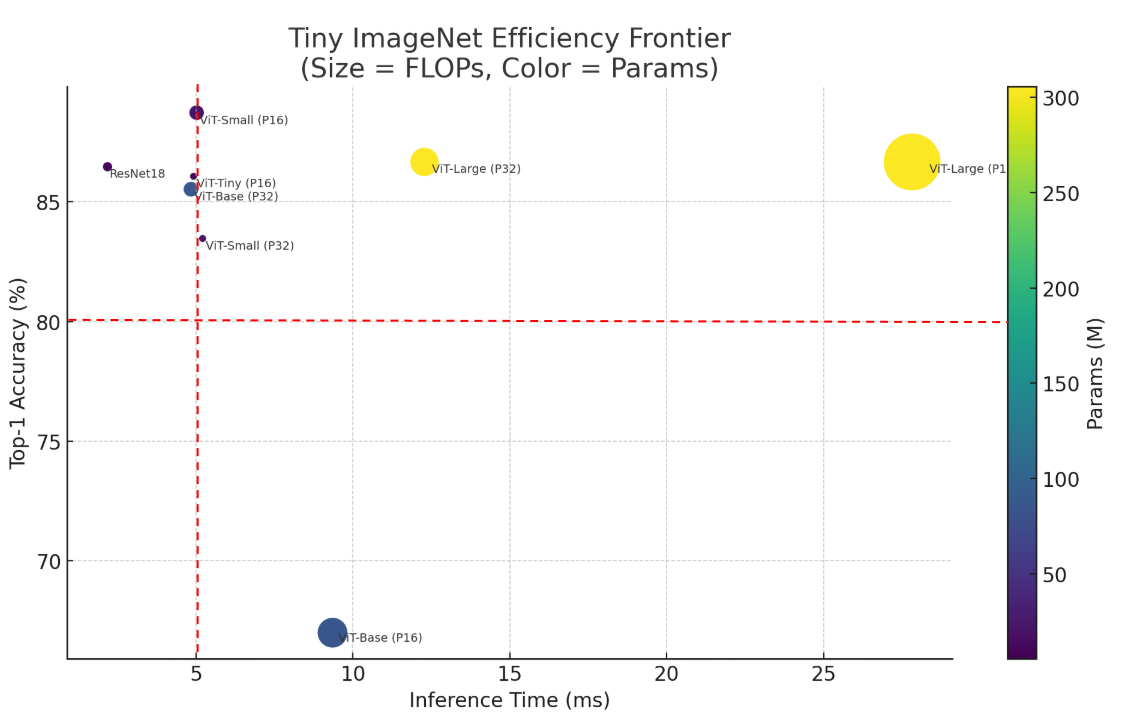}
    \caption{Tiny ImageNet Efficiency Frontier}
    \label{fig:tiny_frontier}
  \end{subfigure}
  \hfill
  \begin{subfigure}[b]{0.48\textwidth}
    \centering
    \includegraphics[width=\textwidth]{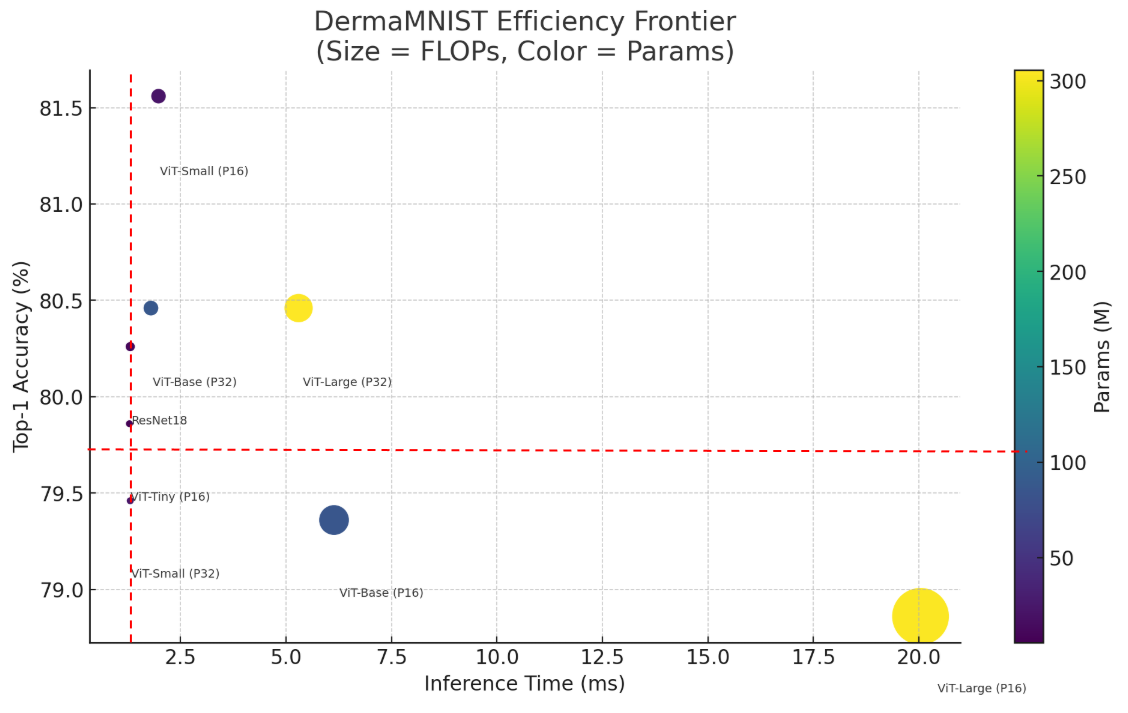}
    \caption{DermaMNIST Efficiency Frontier}
    \label{fig:derma_frontier}
  \end{subfigure}
  \caption{Inference time vs. Accuracy, with marker size/color indicating the number of parameter count for two datasets. The red lines show the minimum amount for accuracy and inference time.}
  \label{fig:efficiency_frontiers}
\end{figure}

Below are the Efficiency Frontier figures (Fig.~\ref{fig:efficiency_frontiers}) that efficiently capture the visual representations of trade-offs we discussed in Tables \ref{tab:tinyimagenet_rel_er} and \ref{tab:dermamnist_rel_er}. 

\section{Discussion and Future Work}

\subsection{Discussion}

Our study evaluated various Vision Transformer (ViT) variants and ResNet18 on two datasets: Tiny ImageNet-200 and DermaMNIST. The key findings are as follows:

\begin{itemize}
    \item \textbf{Accuracy–Efficiency Trade-off:} ViT-Base (Patch 16) achieved the highest accuracy (80\%) on Tiny ImageNet, outperforming the baseline ViT-Base (Patch 32) by approximately 5–7 percentage points. However, this performance came at the cost of significantly increased inference time and model size. In contrast, ViT-Small (Patch 16) offered a balance, maintaining accuracy within ~3 percentage points of ViT-Base (Patch 16) while reducing inference time and model size by approximately 3–4× \cite{b24}.
    \item \textbf{DermaMNIST Performance:} On the DermaMNIST dataset, ViT-Base (Patch 16) again led in accuracy (70\%), with ViT-Small (Patch 16) closely following at 68\%. Notably, ViT-Small (Patch 32) and ViT-Base (Patch 32) underperformed, underscoring the importance of smaller patch sizes for fine-grained medical image classification \cite{b25}.
    \item \textbf{Model Size and FLOPs:} ViT-Small models demonstrated a substantial reduction in parameters and FLOPs compared to ViT-Base models, enhancing computational efficiency without a significant loss in accuracy \cite{b26}.
\end{itemize}

These results highlight the potential of ViT-Small (Patch 16) as an optimal choice for applications requiring a balance between accuracy and efficiency \cite{b27}.

\subsection{Limitations}

Despite promising results, our study has several limitations:

\begin{itemize}
    \item \textbf{Hardware Constraints:} Evaluations were conducted on cloud-based GPUs (e.g., Colab with Tesla T4), which may not accurately reflect real-world performance on edge devices \cite{b28}.
    \item \textbf{Limited Model Spectrum:} Only ViT variants and ResNet18 were considered. Excluding other lightweight models like MobileNetV3, EfficientNetV2, and ConvNeXt may have overlooked potentially more efficient architectures \cite{b29}.
    \item \textbf{Patch Size and ViT Design Bias:} The study focused on fixed ViT designs with patch sizes of 16 and 32, without exploring architectural enhancements such as Shifted Patch Tokenization (SPT) or Local Self-Attention (LSA) \cite{b30}.
    \item \textbf{Absence of Model Compression Techniques:} Techniques like quantization, pruning, and knowledge distillation were not applied, potentially limiting further reductions in inference time and memory usage \cite{b31}.
    \item \textbf{Single-Domain Training:} Models were trained separately on DermaMNIST or Tiny ImageNet, without leveraging multi-task learning or transfer learning to exploit shared representations \cite{b32}.
\end{itemize}

\subsection{Future Work}

To address these limitations and build upon our findings, future research should focus on:

\begin{itemize}
    \item \textbf{Edge/On-Device Evaluation:} Benchmarking ViT-Small (Patch 16) and other promising models on mobile processors (e.g., Snapdragon, Apple M1) and NVIDIA Jetson for real-time applications in medical diagnostics or drone surveillance \cite{b33}.
    \item \textbf{Exploring Alternative Lightweight Architectures:} Evaluating models like MobileNetV3, EfficientNet-Lite, ConvNeXt-Tiny, and MobileViT to identify better accuracy-efficiency trade-offs compared to ViTs \cite{b34}.
    \item \textbf{Architectural Refinements:} Investigating patch size tuning (e.g., 8 or 12), implementing SPT and LSA to enhance generalization on small datasets, and exploring hybrid CNN-ViT fusion for improved spatial and global context modeling \cite{b35}.
    \item \textbf{Model Compression Techniques:} Applying pruning to reduce FLOPs, quantization for low-bit inference, and distillation from ViT-Base (Patch 16) to smaller models to decrease inference time and model size while retaining accuracy \cite{b36}.
    \item \textbf{Cross-Dataset Transfer Learning:} Fine-tuning pre-trained ViT-Small (Patch 16) models across both datasets to improve sample efficiency and generalization, potentially enhancing performance on small, low-resolution datasets \cite{b37}.
\end{itemize}

By addressing these areas, future work can further optimize lightweight models for deployment in resource-constrained environments, advancing the field of computer vision in practical applications

\section{Conclusion}

In this study, we explored the use of Vision Transformers (ViTs) and ResNet18 for skin disease classification using the DermaMNIST dataset and object classification using the Tiny ImageNet dataset. Our objectives were to identify models that strike an optimal balance between accuracy, inference time, and model size, especially in resource-constrained environments such as mobile devices and drones.

The results indicated that ViT-Small (Patch 16) offers an excellent trade-off between accuracy and efficiency. On Tiny ImageNet, ViT-Small (Patch 16) achieved performance within ~3 percentage points of ViT-Base (Patch 16), while being 3-4 times faster in inference and using one-quarter of the parameters. Similarly, on DermaMNIST, ViT-Small (Patch 16) showed minimal performance loss (~2–3 points) compared to ViT-Base (Patch 16), but with significantly lower computational cost. This suggests that ViT-Small (Patch 16) is well-suited for applications that require real-time decision-making, such as medical diagnostics on mobile phones or surveillance on drones.

We also identified some limitations, including the absence of model compression techniques, the lack of transfer learning across tasks, and the focus on only a few ViT variants and ResNet18. Future work should address these limitations by incorporating techniques like pruning, quantization, and knowledge distillation, as well as exploring multi-task learning and transfer learning. Additionally, evaluating models on edge devices such as mobile phones and embedded systems will provide better insights into their real-world deployment potential.

In conclusion, our study demonstrates the feasibility of using lightweight models like ViT-Small (Patch 16) for high-accuracy, low-latency applications, providing valuable contributions to the development of real-time, memory-efficient AI systems in medical and general-purpose image classification tasks.

\end{document}